\documentclass[10pt,twocolumn,letterpaper]{article}
\usepackage{cvpr}              

\usepackage[dvipsnames]{xcolor}

\definecolor{cvprblue}{rgb}{0.21,0.49,0.74}
\usepackage[pagebackref,breaklinks,colorlinks,citecolor=cvprblue]{hyperref}
\usepackage{enumitem}

\usepackage{amsmath}
\usepackage{amsfonts}
\usepackage{amssymb}
\usepackage{epsfig}
\usepackage{graphicx}

\usepackage{booktabs}
\usepackage{multirow}

\newcommand{\MT}[1]{PGSG}

\def\paperID{3882} 
\def\confName{CVPR}
\def\confYear{2024}

\title{From Pixels to Graphs: Open-Vocabulary Scene Graph Generation with Vision-Language Models}

\author{
Rongjie Li\textsuperscript{\rm 1}\thanks{
  Equal contribution and this work was partially done when  Li was a research intern at the Shanghai AI Laboratory.
  ‡ denotes corresponding author.
  Code is available: \url{https://github.com/SHTUPLUS/Pix2Grp_CVPR2024}}
\quad Songyang Zhang\textsuperscript{\rm 2}\footnotemark[1]
\quad Dahua Lin \textsuperscript{\rm 2}
\quad Kai Chen \textsuperscript{\rm 2}\footnotemark[3]
\quad Xuming He\textsuperscript{\rm 1,3}\footnotemark[3] \\
\textsuperscript{\rm 1}School of Information Science and Technology, ShanghaiTech University \quad \\
\textsuperscript{\rm 2}Shanghai AI Laboratory\\
\textsuperscript{\rm 3}Shanghai Engineering Research Center of Intelligent Vision and Imaging\\
\{lirj2, hexm\}@shanghaitech.edu.cn, \{zhangsongyang,lindahua,chenkai\}@pjlab.org.cn,
}

\begin{document}
\maketitle
\begin{abstract}

Scene graph generation (SGG) aims to parse a visual scene into an intermediate graph representation for downstream reasoning tasks.
Despite recent advancements, existing methods struggle to generate scene graphs with novel visual relation concepts.
To address this challenge, we introduce a new open-vocabulary SGG framework based on sequence generation.
Our framework leverages vision-language pre-trained models (VLM) by incorporating an image-to-graph generation paradigm.
Specifically, we generate scene graph sequences via image-to-text generation with VLM and then construct scene graphs from these sequences.
By doing so, we harness the strong capabilities of VLM for open-vocabulary SGG and seamlessly integrate explicit relational modeling for enhancing the VL tasks.
Experimental results demonstrate that our design not only achieves superior performance with an open vocabulary but also enhances downstream vision-language task performance through explicit relation modeling knowledge.

\end{abstract}


\vspace{-0.26cm}
\section{Introduction}\label{sec:intro}

The main objective of scene graph generation (SGG) is to parse an image into a graph representation that describes visual scenes in terms of object entities and their relationship. 
Such a generated scene graph can serve as a structural and interpretable representation of visual scenes, facilitating connections between visual perception and reasoning \cite{wang2018scene}. 
In particular, it has been widely used in various vision-language (VL) tasks, including visual question answering \cite{teney2017graph, shi2019explainable, hildebrandt2020scene, hudson2019learning, hudson2019gqa}, image captioning \cite{yang2019auto, yang2021reformer}, referring expressions \cite{yang2019cross} and image retrieval \cite{johnson2015image}.

\begin{figure}[t]
    \centering
    \includegraphics[width=0.9\linewidth]{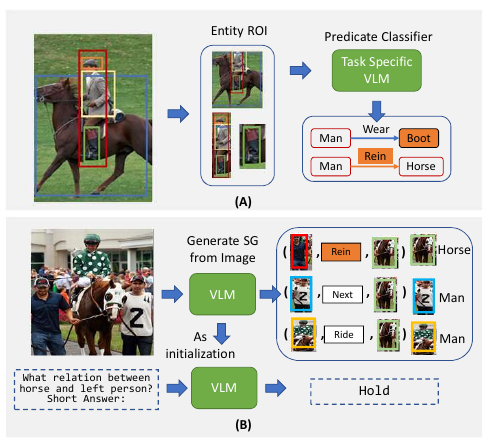}
    \vspace{-0.2cm}
    \caption{\textbf{An illustration of open-vocabulary SGG paradigm comparison.}
    (A) previous work adopt the task-specific VLM as predicate classifiers from given entity proposals;
    (B) Our framework offers a unified framework for generating scene graph with novel predicates from images directly and conducting VL tasks.}
    \label{fig:ad_fig}
    \vspace{-0.35cm}
\end{figure}

Most previous works have focused on addressing the SGG problems in a close-world setting. While much effort has been made in tackling long-tailed data bias \cite{tang_unbiased_2020, guo2021general, zheng2023prototype} and insufficient labeling~\cite{chen_scene_2019, zhang2022fine, li2022nicest}, the typical label space of those methods only covers a limited subset of the diverse visual relationships in the real world.
This can result in incomplete scene representations and also domain gaps when used as intermediate representations in downstream vision-language tasks \cite{wang2022sgeitl}.
To address this limitation, recent works~\cite{he2022towards, zhang2023learning, yu2023visually} start to tackle the SGG problem under various open-vocabulary settings by exploiting the image-text matching capability of pre-trained vision-language models (VLM).
Nevertheless, these attempts typically focus on a simplified setting of the open-vocabulary SGG task, such as allowing only novel entities~\cite{he2022towards, yu2023visually} (Ov-SGG), or a subtask of SGG, such as classifying open-set predicates with given entity pairs~\cite{he2022towards, zhang2023learning}, as shown in part A of Fig.~\ref{fig:ad_fig}.
It remains open to building an end-to-end SGG model in a general open-vocabulary setting. Moreover, those methods often employ an additional pre-training step to enhance the representation power of VLM on relation modeling, which induces a high training cost for large-scale models.

In this work, we aim to address the open-vocabulary SGG problem in a more general setting, i.e., generating scene graphs with both known and novel visual relation triplets from pixels.
To this end, we propose an efficient end-to-end framework that leverages the image-to-text generation paradigm of pre-trained VLMs, as illustrated in part B of Fig.~\ref{fig:ad_fig}.
This framework, dubbed Pixels to Scene Graph Generation with Generative VLM (\MT{}),
In particular, we formulate the SGG as an image-to-sequence problem and introduce a fine-tuning strategy based on generative VLMs.
In contrast to previous methods relying on image-to-text matching VLMs, our generative framework provides a more efficient way to utilize the rich visual-linguistic knowledge of pre-trained VLMs for relation-aware representation without requiring additional pre-training of VLMs. In addition, by converting SGG into a sequence generation problem, our method unifies the SGG task with a diverse set of VL tasks under the generative framework, which enables us to transfer visual relation knowledge to other VL tasks in a seamless manner (e.g., via model initialization).

Specifically, we develop an image-to-graph generation framework consisting of three main components. First, we introduce \textit{scene graph prompts}, which transform scene graphs into a sequence representation with \textit{relation-aware tokens}. Given those graph prompts, we then employ a pretrained VLM to generate a corresponding scene graph sequence for each input image. Finally, we design a plug-and-play \textit{relationship construction module} to extract locations and categories of relation triplets, which produces the output scene graph. Our generation pipeline only requires fine-tuning on target SGG datasets. Such a framework enables us to generate scene graphs with diverse predicate concepts thanks to the large capacity of pre-trained VLMs.
Moreover, our unified framework supports a seamless adaptation strategy, employing our fine-tuned VLM as an initialization for image-to-text generative models in various VL tasks.

We validate our framework on three SGG benchmarks: Panoptic Scene Graph~\cite{yang2022panoptic}, OpenImages-V6~\cite{OpenImages}, and Visual Genome~\cite{krishna2017visual}, achieving state-of-the-art performance in the general open-vocabulary setting.
Furthermore, we apply our SGG-based VLM to multiple VL tasks and obtain consistent improvements, highlighting our effective relational knowledge transfer paradigm.

In summary, the contribution of our work is threefold: 
\begin{itemize}
	\item We propose a novel framework for a general open-vocabulary SGG problem based on the image-to-text generative VLM, which can be seamlessly integrated with other VL tasks.
	\item Our method introduces a scene graph prompt and a plug-and-play relation-aware converter module on top of generative VLM, allowing more efficient model learning. 
	\item Our framework achieves superior performance on the general open-vocabulary SGG benchmarks and consistent improvements on downstream VL tasks. 
\end{itemize}

\vspace{-0.1cm}
\section{Related Work}
\vspace{-0.1cm}
\noindent \textbf{Scene Graph Generation}
Scene graph generation (SGG) aims to localize and classify entities and visualize the relations between them in images. 
The concept of SGG was first introduced in~\cite{krishna2017visual}. Initially, it was used as an auxiliary representation to enhance image retrieval~\cite{johnson2015image, schuster2015generating}. 
The diverse semantic concepts and compositional structures of visual relations create a large concept space, making this task difficult. 
Researchers have approached SGG from various perspectives, including addressing intrinsic long-tailed data bias \cite{chen_knowledge-embedded_2019, tang_unbiased_2020, li2021bipartite, li2022ppdl, li2022devil}, incorporating scene context to model diverse visual concepts \cite{zellers_neural_2017, tang_learning_2018, zareian_bridging_2020, lin2022ru}, and reducing labeling costs through weakly or semi-supervised training \cite{zhong2021learning, yao2021visual, li2022nicest, zhang2023learning}. 
However, most SGG work focuses on limited categories, while few address SGG models' generalization capabilities.
Recent studies, such as \cite{he2022towards}, have systematically investigated this problem by designing VL-pretrained models to classify predicates with unseen object categories, and \cite{zhang2023learning} have investigated visual relations with unseen objects. We generate scene graphs with unseen predicate categories to explore a more realistic setting.

\noindent \textbf{Scene Graph for VL Tasks}
Scene graphs have been extensively studied for improving vision-language (VL) tasks. Early works used scene graphs to improve image retrieval. Several studies showed that scene graphs can be used for image captioning, visual question answering (VQA), and visual grounding. 
Adding explicit scene graphs to VL models for downstream tasks is difficult.
As discussed in \cite{nguyen2021defense, wang2019role}, generated scene graph noise can harm VL models. 
Second, SGG models trained on public datasets with few classes often have a large semantic domain gap compared to visual concepts needed for downstream VL tasks \cite{wang2022sgeitl}. Some works incorporate scene graph representations within model training to address these issues \cite{yao2022pevl, wang2022sgeitl}. Our work aligns with this line: we formulate a unified framework that allows joint optimization between explicit scene graph generation modeling and VL tasks.

\noindent \textbf{Open-vocabulary Scene Graph Generation}
Previous research has predominantly focused on enhancing the generalization of new entity combinations in visual relationships \cite{krishna2017visual,tang_unbiased_2020, kan2021zero, he2022towards, zhang2023learning}. Notably, He et al. \cite{he2022towards} and Yu et al. \cite{yu2023visually} have recently addressed open-vocabulary predicate classification through visual-language pre-training. Nevertheless, these methods struggle to effectively detect visual relations involving unseen predicates in real-world scenarios.
Additionally, some recent works \cite{liao2022gen, ning2023hoiclip} have attempted to handle the challenge of unseen verbs in human-object interaction detection by employing CLIP as a teacher model for feature representation. However, these approaches often neglect to explore the synergies between open-vocabulary SGG and VL tasks.
In this study, we introduce a unified framework leveraging VLMs to tackle open-vocabulary SGG and enhance the reasoning capabilities of VLMs.

\vspace{-0.1cm}
\section{Preliminary}
\vspace{-0.1cm}
\subsection{Scene Graph Generation}\label{subsec:sgg}
\vspace{-0.1cm}

The goal of scene graph generation is to generate a scene graph $\mathcal{G}_{sg}=\{\mathcal{V}_e,\mathcal{R}\}$ from an image, which consists of visual relationships $\mathcal{R}=\{\mathbf{r}_{ij}\}_{i\neq j}$ and $N^v$ entities $\mathcal{V}=\{\mathbf{v}_i\}_{i=1}^{N^v}$. 
The relation triplet $\mathbf{r}_{ij}=(\mathbf{v}_i, c^e_{ij}, \mathbf{v}_j)$ represents the relationship between the $i$-th and $j$-th entities, with a predicate category denoted as $c^e_{ij}$.
The entity $\mathbf{v}_i=\{c^v_i,\mathbf{b}_i\}$ consists of a category label $c^v_i$ in entity category space $\mathcal{O}^v$ and a bounding box $\mathbf{b}_i$, indicating its location in the image.
The predicate category $c^e_{ij}$ belongs to a category space $\mathcal{O}^e$.
In this work, we mainly focus on open-vocabulary predicate SGG settings.
For open-vocabulary predicate SGG, the predicate category space $\mathcal{O}^e$ is divided into: seen base classes  $\mathcal{O}^e_b$, and novel unseen classes $\mathcal{O}^e_n$.
Unlike previous works~\cite{he2022towards,zhang2023learning} tackle detecting relations with entity category $\mathcal{O}^v$ with unseen open-vocabulary set, or classifying novel predicates~\cite{he2022towards, yu2023visually} from given GT entities, we explore a more realistic and challenging setting that generates scene graphs with unseen predicate concepts.

\subsection{Vision-language Models}\label{subsec:vlm}
In this work, we use detector-free BLIP \cite{li2022blip} and instructBLIP as base VLM, which achieves state-of-the-art performance on many VL tasks.
It performs the \textit{image to sequence generation} task with a vision encoder-text decoder architecture.
The vision encoder, denoted as $\mathcal{F}_{v\_enc}$, maps the input image $\mathbf{I}$ to vision features $\mathbf{Z}^v \in \mathbb{R}^{M\times d}$.
For sequence generation, the text decoder with token predictor $\mathcal{F}_{t\_dec}$ outputs hidden states of sequence $\mathbf{H}^s = [\mathbf{h}_0,..., \mathbf{h}_{T'}] \in \mathbb{R}^{T' \times d}$ from encoder features $\mathbf{Z}^v$, by auto-regressive generation methods such as beam search or nucleus sampling.
The token sequence $\mathbf{t}^s = [t_{0},..., t_{T'}] \in \mathbb{D}^{T'}$ with token classification score $\mathbf{P}^s = [\mathbf{p}_{0},..., \mathbf{p}_{T'}] \mathbb{R}^{T' \times |\mathcal{C}_{voc}|}$ in vocabulary space $\mathcal{C}_{voc}$.

\begin{figure*}[h]
    \centering
    \includegraphics[width=0.78\linewidth]{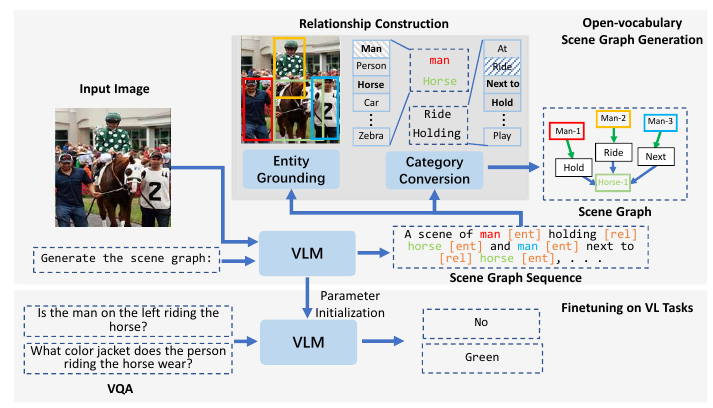}
    \vspace{-0.3cm}
    \caption{\textbf{Illustration of overall pipeline of our \MT{}.} We generate scene graph sequences from the images using VLM. Then, the relation construction module grounds the entities and converts categorical labels from the sequence. For VL tasks, the SGG training provides parameters as initialization for VLM in fine-tuning.
    }
    \label{fig:main_fig}
    \vspace{-0.55cm}
\end{figure*}
 
\vspace{-0.18cm}
\section{Our Approach}
\vspace{-0.1cm}

\vspace{-0.1cm}
\subsection{Method Overview}\label{subsec:overview}
\vspace{-0.1cm}

Our proposed PGSG framework formulates open-vocabulary SGG as an image-to-sequence generation task, which is achieved by a network $\mathcal{F}_{sg}$ with parameters $\mathbf{\Theta}_{sg}$ .
Moreover, with a unified formulation, the \MT{} framework enables a seamless transfer of explicit scene graph representation knowledge by $\mathbf{\Theta}_{sg}$ to VL tasks.
Specifically, our framework is composed of three components: scene graph sequence generation, relation triplet construction, and adaptation to downstream VL tasks, as shown in Fig. \ref{fig:main_fig}.
First, we generate the scene graph sequence $\mathbf{s}_{sg}$ from the input image $\mathbf{I}$ by image-to-text generation of VLM with the scene graph sequence prompts (Sec.~\ref{subsec:sggp}).
Second, the relationship construction module extracts $\mathcal{R}$ from the scene graph sequence $\mathbf{s}_{sg}$ to construct the scene graph $\mathcal{G}_{sg}$ (Sec.~\ref{subsec:ist_cvt}).
In addition, we introduce the learning and inference pipeline of our SGG framework in Sec.~\ref{subsec:le_in}.
Finally, we adapt the VLM with $\mathbf{\Theta}_{sg}$ for transferring the explicit scene graph representation knowledge to VL tasks (Sec.~\ref{subsec:downstm}).

\vspace{-0.1cm}
\subsection{Scene Graph Sequence Generation}\label{subsec:sggp}
\vspace{-0.1cm}
The scene graph sequence $\mathbf{s}_{sg}$ is composed of a set of relation triplets $\mathcal{R}$ from $\mathcal{G}_{sg}$.
To achieve this, we propose a scene graph sequence prompt:

`` \textit{Generate the scene graph of [triplet sequence] and [triplet sequence] ...} ''

\noindent It consists of two primary components: the prefix instruction "Generate the scene graph of" and $K$ relation triplet sequences, separated by ``and'' or ``,''.
Specifically, each triplet sequence is transformed from the $\mathbf{r}_{ij}$ by the natural language grammar of subject-predicate-object:

`` \textit{$\mathbf{t}^v_i$ [ENT] $\mathbf{t}^e_{ij}$ [REL]  $\mathbf{t}^v_j$ [ENT]} '' 

\noindent The token sequences $\mathbf{t}^v_i$, $\mathbf{t}^v_i$, and $\mathbf{t}^e_{ij}$ represent the tokenized category names of the subject, object entities $c^v_i, c^v_j$, and predicate $c^e_{ij}$.
We also introduce specified relation-aware tokens, [ENT] and [REL], to represent the compositional structure of relationships and the position of entities.

The text decoder of VLM takes $\mathbf{Z}^v$ and the prefix instruction as input and generates $\mathbf{s}_{sg}$. This procedure is similar to standard image captioning.
By using the vocabulary space of natural language, we can use generalized semantic representation to generate the visual relationship.
The following modules extract both the spatial and categorical information and construct the instance-aware relation triplet from the sequence using relation-aware tokens.

\vspace{-0.1cm}
\subsection{Relationship Construction}\label{subsec:ist_cvt}
\vspace{-0.1cm}
As referred to in Sec. \ref{subsec:sgg}, the standard visual relation triplet $r_{ij}$ contains the predicate category label $c^e_{ij}$, entities with position $\textbf{b}_i, \textbf{b}_j$, and category $c^v_i, c^v_j$.
The relation construction aims to extract spatial and category labels from $\mathbf{s}_{sg}$ to construct the relation triplets.
This relationship construction has two submodules: 1) \textit{Entity Grounding Module}, which outputs entity positions, and 2) \textit{Category Convert Module}, which converts language prediction from vocabulary space to category space.

\vspace{-0.22cm}
\subsubsection{Entity Grounding Module}\label{subsec:pos_cvt}
\vspace{-0.1cm}

The \textit{entity grounding module} predicts the bounding box of an entity to ground the generated relations within the scene graph sequence.
Unlike existing multi-output VLMs~\cite{yao2022pevl, wang2022ofa, yang2022unitab} that output sequences comprising a mixture of coordinates and words, the entity grounding module predicts bounding boxes $\mathbf{B} \in \mathbb{R}^{2N \times 4}$ of entity token sequence ``$\mathbf{t}^v_i$ [ENT]'', ``$\mathbf{t}^v_j$ [ENT] '' from $N$ relation triplets of $\mathbf{s}_{sg}$.
This process significantly enhances the quality of modeling spatial coordinates for these entities.

\begin{figure*}[h]
    \centering
    \includegraphics[width=0.95\linewidth]{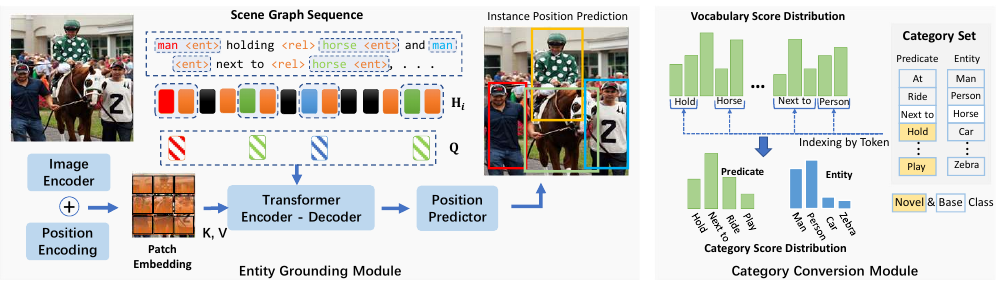}
    \vspace{-0.35cm}
    \caption{\textbf{Illustration of entity grounding module and category conversion module.} 
     Left): The entity grounding module localizes the entities within scene graph sequences by predicting their bounding boxes.
    Right): The category conversion module maps the vocabulary sequence prediction into categorical prediction.
    }
    \label{fig:pos_adpter}
    \vspace{-0.65cm}
\end{figure*}

As shown in Fig.~\ref{fig:pos_adpter}, for each token sequence ``$\mathbf{t}^v_i$ [ENT]'', we extract the $\mathbf{b}_i$ by hidden states $\mathbf{H}_i = [\mathbf{h}_{1}^{t_i^v},...,\mathbf{h}_{T_v}^{[ENT]}] \in \mathbf{R}^{T_v \times d}$ of , where $T_v$ is length of $\mathbf{t}^v_i$.
We utilize the token hidden states as queries $\mathbf{Q}$ to re-attend the image features $\mathbf{Z}_v$ through the attention mechanism to more accurately locate the objects.
We first transform $[\mathbf{h}_{1}^{t_i^v},...,\mathbf{h}_{T_v}^{[ENT]}] $ into query vector $\mathbf{q}^{ent}_{i} \in \mathbf{R}^{d_q} $ by average pooling and linear projection:
\vspace{-0.3cm}
\begin{align}
    \mathbf{q}^{ent}_i = \frac{1}{T_v} \sum_{k=1}^{T_v} \textbf{h}_k^{t_i^v} \cdot \mathbf{W}^{\text{T}}_q.
    \vspace{-0.6cm}
\end{align}
Subsequently, we decode the $\mathbf{b}_i$ by cross-attention between queries of $2N$ entity token sequences: $\mathbf{Q} = \{\mathbf{q}^{ent}_{1},..., \mathbf{q}^{ent}_{2N}\} \in \mathbb{R}^{2N \times d} $ and $\mathbf{Z}_v$ using transformer encoder $\mathcal{F}_{enc}(\cdot)$ and decoder $\mathcal{F}_{dec}(\cdot)$.
Finally, the $\mathbf{B}$ is predicted by the feed-forward network $\text{FFN}(\cdot)$.
\vspace{-0.15cm}
\begin{align}
    \hat{\mathbf{Q}}&=\mathcal{F}_{dec}(\mathbf{Q}, \mathcal{F}_{enc}(\mathbf{Z}_v)),\\
    \mathbf{B}&=\text{FFN}(\hat{\mathbf{Q}}).
    \vspace{-0.5cm}
\end{align}\vspace{-0.1cm}
By introducing this dedicated module, we are able to generate standard scene graphs with instance-level spatial prediction from an image-level scene graph sequence.

\vspace{-0.25cm}
\subsubsection{Category Conversion Module}\label{subsec:voc_cvt}
\vspace{-0.15cm}
During the sequence generation, the category of entities and predicates is represented by a token from the language vocabulary space $\mathcal{C}_{voc}$.
However, the SGG benchmark category concept space is only part of the language vocabulary.
To align the open-vocabulary predictions with the category space of the target SGG benchmark, we need to convert the vocabulary token into a category label.
As illustrated in Fig.~\ref{fig:pos_adpter}, the \textit{Category Conversion Module} maps the tokens prediction score from the vocabulary space $\mathcal{C}_{voc}$ to the entity and predicate category space $\mathcal{O}^v, \mathcal{O}^e$ in the target SGG benchmark, respectively.

The category conversion module is parameter-free, which transforms the vocabulary score $\mathbf{P}^s_{i}, \mathbf{P}^s_{ij} \in \mathbb{R}^{T_v \times |\mathcal{C}_{voc}|}$ of each entity and the predicate token sequences $\mathbf{t}^v_i$ and $\mathbf{t}^e_{ij}$ into the categorical prediction scores $\mathbf{p}^v_i \in \mathbb{R}^{|O_v|}$ and $\mathbf{p}^e_{ij} \in \mathbb{R}^{|O_e|}$ respectively.
To achieve this transfer, we first tokenize target category sets, include entity $\mathcal{O}^v$, and predicate $\mathcal{O}^e$ into the token sequences $\mathbf{t}_c^e$ and $\mathbf{t}_c^v$ by tokenizer of VLM.
\vspace{-0.09cm}
\begin{align}
    \mathbf{t}_c^e = [t_{1}^1, t_{2}^1,t_{1}^2, ... t^{C_e}_{T_{C_e}} ],~~
    \mathbf{t}_c^v = [t_{1}^1, t_{2}^1,t_{1}^2, ... t^{C_v}_{T_{C_v}} ] .
    \vspace{-0.3cm}
\end{align}
Here, the superscript $c$ denotes the category index, and $T$ is the sequence length for each tokenized category name.
Subsequently, we compose $\mathbf{p}^e_r$ and $\mathbf{p}^v_i$ from $\mathbf{p}^s$ using the corresponding token indices of $\mathbf{t}_c^v$ and $\mathbf{t}_c^e$:
\vspace{-0.19cm}
\begin{align}
    \mathbf{p}^e_{ij}=\{\frac{\beta_{1}}{2}\sum_{i=1}^{2} \mathbf{P}^s_{ij}[t_i^1];~...~;~\frac{\beta_{C_e}}{T_{C_e}} \sum_{i=1}^{T_{C_e}} \mathbf{P}^s_{ij}[t_i^{C_e}]\}.
\vspace{-0.4cm}
\end{align}
Here, $[\cdot]$ represents the indexing operation, and $\{\cdot;\cdot\}$ denotes the concatenation operation. The same procedure is applied to the scores for entity classification $\mathbf{p}^v_i$.
Additionally, if the generated tokens $\mathbf{t}^v_i$ and $\mathbf{t}^e_{ij}$ exactly match the given category name, we amplify the score of this category by $\beta_i$.
To this end, the category conversion module enables the evaluation and analysis of SGG performance with respect to the defined categories.

\vspace{-0.1cm}
\subsection{Learning and Inference}\label{subsec:le_in}
\vspace{-0.1cm}
\subsubsection{Learning}
\vspace{-0.1cm}
For SGG training, we use a multi-task loss consisting of two parts: the loss for standard next token prediction language modeling $\mathcal{L}^{lm}$ and the loss for the entity grounding module $\mathcal{L}^{pos}$. 
The $\mathcal{L}^{lm}$ loss follows the same definition as in the VLM backbone, which is a standard self-regressive language modeling loss. It predicts the next token $t_i$ based on previously generated tokens ($\mathbf{t} = [t_1, t_2,..., t_{i-1}]$) and visual features $\mathbf{Z}^v$. 
This prediction is represented as: $t_i = \mathcal{F}_{vlm}( \mathbf{Z}^v,\mathbf{t}; \mathbf{\Theta}_{sg})$. To train this VLM, we optimize its parameters ($\mathbf{\Theta}_{sg}$) by maximizing the likelihood of correctly predicting each subsequent token in the sequence sequence of length $K$ : $ \sum_{i=1}^{K} \log P(t_i| \mathbf{Z}, t_1, t_2,..., t_{i-1}; \mathbf{\Theta}_{vlm})$. Compared to previous SGG methods, our self-regression-based approach effectively captures the semantic relationships between predicates and entities.
The loss $\mathcal{L}^{pos}$ is for the bounding box regression of  entity grounding module.
For each box prediction $\mathbf{B}$ of scene graph sequence, we have GIOU and L1 loss for calculating the distance $\mathbf{B}$ and GT $\mathbf{B}_{gt}$: $\mathcal{L}^{pos} = \text{GIOU}(\mathbf{B}, \mathbf{B}_{gt}) + || \mathbf{B} - \mathbf{B}_{gt}||_1$.

\vspace{-0.28cm}
\subsubsection{Inference}
\vspace{-0.1cm}
During inference, we first generate the scene graph sequence using VLMs.
The following relationship construction module extracts category labels and spatial positions for constructing visual relations from the sequences. 

\noindent \textbf{Scene Graph Sequence Generation}
The scene graph sequence is generated using the image-to-text generation of VLMs.
Our goal is to generate a diverse set of relationships to create a more comprehensive scene graph representation.
To achieve this, we employ the nucleus sampling strategy~\cite{holtzman2019curious} with multiple round sequence generation.
Specifically, for each image, we generate M sequences, whose maximum length is L.
This strategy encourages the model to generate diversely, allowing scene graph predictions to capture various visual entities and their relationships.

\noindent \textbf{Relationship Triplet Construction}
To construct relationship triplets $\hat{\mathbf{r}}_{ij} = (\mathbf{v}_i, c^e_{ij}, \mathbf{v}_j)$ from the scene graph sequence, we initially extract the token sequence of subject and object entities $\mathbf{t}^v_i$, $\mathbf{t}^v_j$, and predicate $\mathbf{t}^e_{ij}$.
Specifically, we employ a heuristic rule to match patterns such as ``\textit{subject [ENT] predicate [REL] object [ENT]}'', allowing us to extract these triplets.
Subsequently, utilizing a relationship construction module, we extract $\mathbf{b}_i, \mathbf{p}^v_{i}$ of entities $\mathbf{v}_i$,and $\mathbf{b}_j, \mathbf{p}^v_{j}$ for $\mathbf{v}_j$, with predicate scores $\mathbf{p}^v_{ij}$.
To obtain the category labels of triplets $c^v_i, c^v_j, c^e_{ij}$, we select the top 3 categories according to prediction scores: $\mathbf{p}^v_{i}$, $\mathbf{p}^v_{j}$, and $\mathbf{p}^v_{ij}$.

\noindent \textbf{Post-Processing}
We refine the initial triplets to obtain the final relationship triplets through a process of filtering and ranking. 
Firstly, we remove self-connected relationships where the subject and object entities are the same. 
Then, we apply a non-maximum suppression (NMS) strategy to eliminate redundant relationships, resulting in a more concise predicted scene graph. Finally, we rank all relationships based on the score of each component triplet.
This score, denoted as $S^t$, is calculated as the product of prediction scores for the involved entities $s^v_i, s^v_j$ and the predicate $s^e_p$, extracted from category score distributions $\mathbf{p}_i^v, \mathbf{p}_j^v, \mathbf{p}_{ij}^e$ using their respective category indices.

\vspace{-0.1cm}
\subsection{Adaptation to Downstream VL Task}\label{subsec:downstm}
\vspace{-0.1cm}

In our framework, depicted in Fig.~\ref{fig:main_fig}, we employ a unified image-to-text generation approach, facilitating seamless knowledge transfer for relation modeling.
We initialize the visual encoder $\mathbf{\Theta}_{v\_enc}$ and text decoder $\mathbf{\Theta}_{t\_dec}$ parameters of the Scene Graph Generation (SGG) model, denoted as $\mathbf{\Theta}_{sg}$, for VLMs fine-tuning on VL tasks. 
Additionally, we employ initial pre-trained weights for the token predictors of text decoders to maintain vocabulary prediction quality during fine-tuning. Modules of the VLM not utilized during SGG training, such as the text encoder, retain their initial pre-trained weights. This adaptive knowledge sharing from SGG parameters during fine-tuning enhances the performance of VL tasks.

\vspace{-0.25cm}
\section{Experiments}
\vspace{-0.1cm}

\subsection{Experiments Configuration}
\vspace{-0.1cm}
We evaluate our method on both SGG and downstream VL tasks.
For SGG benchmarks, we evaluate \MT{} on three benchmarks: Panoptic Scene Graph Generation (PSG)~\cite{yang2022panoptic}, Visual Genome (VG)~\cite{krishna2017visual}, and OpenImage V6 (OIv6)~\cite{OpenImages}.
We randomly select 50\% predicate categories as novel classes for open-vocabulary predicate SGG.

We assess our method's performance in the SGG task using three evaluation protocols: SGDet, PCls, and SGCls, as proposed by~\cite{xu_scene_2017}. We calculate overall recall (R@K) and class-balance metric mean recall (mR@K) on the top K predictions. 
Furthermore, we present mR@K specifically for novel classes to illustrate the model's performance in generalizing to unseen predicates.

For VL tasks, we inspect our model on
visual grounding on RefCOCO/+/g~\cite{yu2016modeling, mao2016generation},
visual question answering on GQA~\cite{hudson2019gqa}
and image captioning on COCO~\cite{chen2015microsoft}.

We use the BLIP \cite{li2022blip} as our VLM backbone, which employs ViT-B/16 as the visual backbone and BERT$_{base}$ as the text decoder.
For downstream VL tasks, we initialize the parameters with the pre-trained SGG model.
More implementation details can be found in the supplementary materials.

\begin{table}[t]
    \begin{center}
        \resizebox{0.48\textwidth}{!}{ 
            \begin{tabular}{l|l|l|cc|c}
                \toprule
                \multirow{2}{*}{\textbf{D}}    & \multirow{2}{*}{\textbf{S}} & \multirow{2}{*}{\textbf{M}} & \multicolumn{2}{c|}{\textbf{Novel+base}}        & \textbf{Novel}            \\
                                      &                          &                 & \textbf{mR50/100}  & \textbf{R50/100}   & \textbf{mR50/100}         \\ \midrule
                \multirow{8}{*}{VG}   & \multirow{3}{*}{PCls} & CaCao              & 10.3 / 12.6        & -                   & -                \\
                                      &                          & SVRP             & 8.3 / 10.8         & \textbf{33.5} / \textbf{35.9}           & -                \\ 
                                      &                          & \textbf{\MT{}}* & \textbf{10.8} / \textbf{13.9}        & 26.9 / 33.9         & 5.2 / 7.7               \\  \cmidrule{2-6} 
                                      & \multirow{2}{*}{SGCls}   & SVRP               & 3.2 / 4.5          & 19.1 / 21.5           & -                \\
                                      &                          & \textbf{\MT{}}* &  \textbf{8.4} / \textbf{11.0}         &   \textbf{22.6} / \textbf{27.2}          &  4.8 / 6.0                \\ \cmidrule{2-6} 
                                      &   \multirow{8}{*}{SGDet}  & VS3                & 5.1 / 5.7                & 11.0 / 12.8                  & 0.0 / 0.0               \\
                                      & & SGTR$^\dagger$         & 3.5 / 5.4          & 12.6 / 18.2         & 0.0 / 0.0         \\
                                      &                          & \textbf{\MT{}} & \textbf{6.2 / 8.3} & \textbf{15.8 / 19.1} & \textbf{3.7 / 5.2} \\ \cmidrule{1-1} \cmidrule{3-6} 
                \multirow{2}{*}{PSG}  &    & SGTR$^\dagger$      & 6.4 / 8.4          & 14.2 / 18.2           & 0.0 / 0.0          \\
                                      &                          & \textbf{\MT{}} & \textbf{13.5} / \textbf{16.4} & \textbf{18.0} / \textbf{20.2} & \textbf{7.4} / \textbf{11.3}          \\ \cmidrule{1-1} \cmidrule{3-6} 
                                      
                \multirow{3}{*}{OIv6} &   & SVRP               & -                & -                   & -                \\
                                      &                          & SGTR$^\dagger$                & 11.0 / 16.7          & 36.1 / 38.4           & 0.0 / 0.0          \\
                                      &                          & \textbf{\MT{}} & \textbf{20.8} / \textbf{23.0} & \textbf{41.3} / \textbf{43.3} & \textbf{3.8} / \textbf{8.9}          \\ \bottomrule
            \end{tabular}
        }
    \end{center}
    \vspace{-0.5cm}
    \caption{\textbf{The open-vocabulary scene graph generation on VG, PSG, and OIV6 datasets}
    $\dagger$ denotes the method is reproduced with BLIP visual encoder ViT as the backbone;
    D denotes the dataset name; S is the SGG setting; M represents the model;
    * denotes that we use the grounding truth assignment on predictions. }
    \vspace{-0.65cm}
     \label{tab:ov_sgg}

\end{table}

\vspace{-0.15cm}
\subsection{Open-vocabulary SGG}
\vspace{-0.1cm}

\noindent \textbf{Setup}
We evaluate our design on the VG, PSG, and OIv6 datasets by comparing it with OV-SGG methods: CaCao\cite{yu2023visually}, SVRP\cite{he2022towards}, and VS3\cite{zhang2023learning} in Tab~\ref{tab:ov_sgg}.
The CaCao and SVRP have the capability for open-vocabulary predicate classification with GT entities.
In contrast, the VS3 achieves SGDet on close-set predicate, where relation triplets have unseen entity categories.
Additionally, we adapt a one-stage SGG method (SGTR~\cite{li2022sgtr}) for predictive open-vocabulary SGG since it has the most balanced performance in a close-vocabulary SGG setting.
To apply close-set SGG methods to predicate open vocabulary, we train models on base categories only and set the classifier weights of novel categories to zero at inference time.

\noindent \textbf{VG}
Our method demonstrates strong performance compared to previous methods on PCls, SGCls, and SGDet settings on VG.
Since the VLM cannot take a bounding box as input, we simulate entity-based relations in PCls and SGCls settings by directly replacing entity prediction with GT entities.
Our method improves base+novel class mR@100 and R@100 by 6.5 and 5.7 over SVRP in SGCls.
We outperform SVRP's task-specific pretrain model with a general pretrain VLM in an image-to-text generation fashion.

For the SGDet setting, we compare with two baselines: VS3 and SGTR.
Our method outperforms both methods, achieving 2.9 and 0.9 on mR@100 and R@100 for all categories.
It's worth noting that VS3 and SGTR are unable to generalize to novel classes and can only predict relationships for base categories.

\noindent \textbf{PSG}
Our \MT{} method achieves significant performance gains for both seen base and unseen novel categories on the PSG dataset.
We observe improvements of 9.4 and 14.3 on mR@100 and R@100 compared with SGTR equipped with a pre-trained ViT backbone in the whole category set, respectively.
For novel predicate categories, \MT{} outperforms the strong baseline SGTR method by 4.5 on mR@100.

\noindent \textbf{OpenImage}
Our approach excels on the OpenImage V6 benchmark (see Tab.~\ref{tab:ov_sgg}). Comparing our design to the SGTR baseline, we achieve better mR@100 and R@100 with margins of 5.3 and 4.9.

\vspace{-0.10cm}
\subsection{Close-vocabulary SGG}
\vspace{-0.1cm}
Our method is also tested in a closed-vocabulary SGG.
Our method generates relationships using VLMs without explicitly pairing entity proposals, so we compare it to one-stage SGG methods with top-down architecture, such as PSGTR~\cite{yang2022panoptic}, PSGFormer~\cite{yang2022panoptic}, SGTR~\cite{li2022sgtr}, RelTR~\cite{cong2023reltr}, ISG~\cite{khandelwal2022iterative}, SSRCNN~\cite{teng2021structured}.
In zero-shot triplet SGG, our method shows good compositional generalization.
To ensure a fair comparison, we substituted the category conversion module for a close-set classifier, aligning with previous SGG methods. 
As shown in Tab.~\ref{tab:stdsgg_psg_vg}, the BGSG with the close-set classifier achieves competitive or superior performance when compared to SGG SOTA methods, SGTR, in a standard SGG setting.

\begin{table}[t]
    \begin{center}
        \resizebox{0.49\textwidth}{!}{ 
            \begin{tabular}{c|l|l|c|cc}
                \toprule
                \multirow{2}{*}{\textbf{D}}   & \multirow{2}{*}{\textbf{B}}          & \multirow{2}{*}{\textbf{M}} & \textbf{Zs-SGG} & \multicolumn{2}{c}{\textbf{Close-SGG}} \\
                                     &                             &                    & zR50/100       & mR50/100        & R50/100        \\ \midrule
                \multirow{6}{*}{PSG} & \multirow{3}{*}{R101}       & PSGTR              & 3.1/\textbf{6.4}        & 20.3/21.5       & 32.1/35.3      \\
                                     &                             & PSGFormer          & 2.2/4.9        & 19.3/19.7       & 20.4/20.7      \\
                                     &                             & SGTR               & \textbf{4.1}/5.8        & \textbf{24.3}/\textbf{27.2}       & \textbf{33.1}/\textbf{36.3}      \\ \cmidrule{2-6} 
                                     & \multirow{3}{*}{ViT-B*} & SGTR               & 3.1/3.6        & \textbf{22.1}/\textbf{23.5}       & 30.1/32.0      \\
                                     &                             & \MT{}               & 5.1/7.5        & 14.9/18.1       & 26.0/28.9      \\ 
                                     &                             & \MT{}-c             & \textbf{6.8}/\textbf{8.9}         & 20.9/22.1          & \textbf{32.7}/\textbf{33.4}  \\ \midrule
                \multirow{10}{*}{VG} & X101-FPN                    & SSRCNN             & 3.1/4.5        & 18.6/22.5       & 23.7/27.3      \\ \cmidrule{2-6} 
                                     & R50                         & SVRP               & -            & 10.5/12.8       & 31.8/35.8      \\ \cmidrule{2-6} 
                                     & Swin-L                      & VS3                & -            & -             & 34.5/39.2      \\ \cmidrule{2-6} 
                                     & \multirow{3}{*}{R101}      & RelTR              & 2.6/3.4        & 6.8/10.8        & 21.2/27.5      \\
                                     &                             & ISG                & \textbf{2.7}/3.8        & 8.0/8.8       & \textbf{29.7}/\textbf{32.1}     \\
                                     &                             & SGTR               & 2.5/\textbf{5.8}        & \textbf{12.0}/\textbf{15.2}       & 24.6/28.4      \\ \cmidrule{2-6} 
                                     & \multirow{3}{*}{ViT-B*} & SGTR               & 2.1/4.9        & 8.7/10.1        & 18.1/20.4      \\
                                     &                         & \MT{}              & \textbf{6.2}/\textbf{8.5}        & 8.9/11.5        & 16.7/21.2  \\
                                     &                         & \MT{}-c            & 4.8/7.6         & \textbf{10.5}/\textbf{12.7}          & \textbf{20.3}/\textbf{23.6}             \\ \bottomrule
            \end{tabular}
        } 
    \end{center}
    \vspace{-0.60cm}
    \caption{\textbf{The close-vocabulary SGDet performance on VG and PSG dataset.} 
    D: dataset name; B: visual backbone; M: Method;
    ViT-B*: the use BLIP ViT visual encoder as backbone.
    -c: denotes use the close-set classifier.
    } \label{tab:stdsgg_psg_vg}
    \vspace{-0.67cm}

\end{table}

\noindent \textbf{Zero-shot Triplets}
In the middle part of Tab.~\ref{tab:stdsgg_psg_vg}, we evaluate our method's generalization ability through zero-shot triplet SGG.
On the PSG dataset, our \MT{} method enhances zR@100 by 4.1 over the ViT SGTR baseline and matches the ResNet-101 PSGTR model, despite lower image input.
\MT{} performed well on the VG dataset, outperforming SSRCNN by 4.0 on zR@100 and achieving a 3.6 margin over the ViT SGTR baseline.

\noindent \textbf{Discussion}
We also evaluate our method against previous SGG approaches under standard close-vocabulary SGG settings in the Tab.~\ref{tab:stdsgg_psg_vg}. 
While our method performs similarly to prior works using the same ViT backbone in closed-vocabulary settings, there remains a gap between our results and previous SOTA performances.

\noindent $\bullet$ Limited input resolution: Transformer-based VLMs like BLIP use smaller images (384x384), while traditional SGG methods (e.g., ResNet-101 FPN, 800x1333) may miss or mislocalize small or ambiguous objects, which have been shown to pose challenges in SGG tasks~\cite{li2022sgtr, li2022nicest}.

\noindent $\bullet$ Single-stage training: In \MT{}, the BLIP is used for SGG training directly, as opposed to the conventional two-stage training. 
However, this approach imposes limitations on the model's ability to accurately detect small objects.

\vspace{-0.10cm}
\subsection{Ablation Study}
\vspace{-0.1cm}
We conduct an ablation study to investigate the impact of different model structures and the effectiveness of SGG training.
Specifically, we compare the design choices regarding the entity grounding module with the scene graph prompt on the PSG dataset, which is presented in Tab. \ref{tab:Ablation}.

\noindent \textbf{Entity Grounding Module}
First, we evaluate the effectiveness of \textit{Entity Grounding with Relation-aware Tokens} for predicting entity location in the lower part of Tab.~\ref{tab:Ablation}.
Our approach demonstrates notable improvements in entity detection quality compared to previous methods such as UniTab and OFA, which rely on sequence generation by VLMs incorporating both coordinates and text.
Moreover, we investigate the impact of the number of transformer layers ($L$) in the entity grounding module.
As seen in the upper part of Tab.~\ref{tab:Ablation}, we vary $L$ from 0 to 12. Here, $L=0$ signifies direct prediction of the bounding box from the grounding entity query $\mathbf{Q}$, with a performance plateau at $L=6$.

\noindent \textbf{Effectiveness of SGG Supervision}
To evaluate the impact of scene graph supervision on downstream VL tasks, we fine-tune the BLIP model after training it on entity-only sequence prompts with the same format:``A photo of $\text{entity}_1$ [ENT] and $\text{entity}_2$ [ENT]...''. Fine-tuning is performed on the GQA dataset.
Results from Tab. \ref{tab:GQA} show that positional supervision does not lead to significant improvements over initial pre-training.
In contrast, scene graph supervision enhances the performance of GQA, especially for relation and attribute questions.

\begin{table}[]
    \begin{center}
    \resizebox{0.44\textwidth}{!}{ 
        \begin{tabular}{lcccccc}
            \toprule
            \multicolumn{7}{c}{\textbf{Layer of Entity Grounding Module}}                                                        \\ \midrule
            \multicolumn{1}{l|}{\textbf{L}}       & \textbf{mR20} & \textbf{mR50} & \multicolumn{1}{c|}{\textbf{mR100}} & \textbf{R20}  & \textbf{R50}  & \textbf{R100} \\ \midrule
            \multicolumn{1}{l|}{0}       & 2.0  & 3.8  & \multicolumn{1}{c|}{10.3}  & 7.1 & 13.2 & 20.1 \\
            \multicolumn{1}{l|}{3}       & 3.2  & 8.1  & \multicolumn{1}{c|}{15.9}  & 11.0 & 18.7 & 27.9 \\
            \multicolumn{1}{l|}{6}       & 3.0  & \textbf{8.4 } & \multicolumn{1}{c|}{\textbf{17.0}}  & 11.1 & \textbf{19.2} & \textbf{28.4} \\
            \multicolumn{1}{l|}{12}      & \textbf{3.2}  & 8.1  & \multicolumn{1}{c|}{16.7}  & \textbf{12.1} & 18.9 & 28.2 \\ \midrule
            \multicolumn{7}{c}{\textbf{Relation-aware Token for Entity Grounding}}         \\ \midrule
            \multicolumn{1}{l|}{\textbf{C}}      & \textbf{mR20} & \textbf{mR50} & \multicolumn{1}{c|}{\textbf{mR100}} & \textbf{R20}  & \textbf{R50}  & \textbf{R100} \\ \midrule
            \multicolumn{1}{l|}{\textbf{mixture}}  & 1.2  & 3.1  & \multicolumn{1}{c|}{8.5}   & 6.4  & 11.8 & 16.3 \\
            \multicolumn{1}{l|}{\textbf{tokens}} & \textbf{3.0}  & \textbf{8.4}  & \multicolumn{1}{c|}{\textbf{17.0}}  & \textbf{11.1} & \textbf{19.2} & \textbf{28.4} \\ \bottomrule
        \end{tabular}
    }
    \end{center}
    \vspace{-0.49cm}
    \caption{\textbf{Ablation study for entity grounding module on PSG dataset.}  
    \textbf{L}: indicates the layer of transformer within entity grounding module. 
    \textbf{C}: denotes the different design choices for entity grounding.  
    } 
    \vspace{-0.4cm}
    \label{tab:Ablation}
\end{table}

\begin{table}[]
    \begin{center}
    \resizebox{0.49\textwidth}{!}{ 
        \begin{tabular}{lccccc}
            \hline
            \multicolumn{1}{l|}{\multirow{3}{*}{\textbf{SL}}} & \multicolumn{3}{c|}{\textbf{Open Vocab SGG}}                                     & \multicolumn{1}{c|}{\multirow{2}{*}{\textbf{ZS Trp.}}} & \multirow{2}{*}{\textbf{Time}} \\
            \multicolumn{1}{l|}{}                            & \multicolumn{2}{c|}{\textbf{Novel+Base}}   & \multicolumn{1}{c|}{\textbf{Novel}} & \multicolumn{1}{c|}{}              &                                \\
            \multicolumn{1}{l|}{} & \multicolumn{1}{c}{\textbf{R50}/\textbf{100}}  & \multicolumn{1}{c|}{\textbf{mR50}/\textbf{100}}      & \multicolumn{1}{c|}{\textbf{mR50}/\textbf{100}}     & \multicolumn{1}{c|}{\textbf{R50}/\textbf{100}}      & \textbf{Sec} \\ \midrule
            \multicolumn{1}{l|}{1024}                        & 16.9/18.8  & \multicolumn{1}{c|}{22.9/25.0} & \multicolumn{1}{c|}{6.7/9.9}        & \multicolumn{1}{c|}{3.6/6.4}      & 6.9                            \\
            \multicolumn{1}{l|}{768}                         & 16.8/18.0 & \multicolumn{1}{c|}{23.6/24.9} & \multicolumn{1}{c|}{7.9/10.0}        & \multicolumn{1}{c|}{8.5/8.1}      & 4.8                            \\
            \multicolumn{1}{l|}{512}                         & 15.3/15.6 & \multicolumn{1}{c|}{22.0/22.5} & \multicolumn{1}{c|}{7.4/8.1}        & \multicolumn{1}{c|}{3.5/4.6}       & 2.2                            \\
            \multicolumn{1}{l|}{256}                         & 12.8/12.8 & \multicolumn{1}{c|}{18.3/18.3} & \multicolumn{1}{c|}{5.3/5.3}        & \multicolumn{1}{c|}{2.4/2.4}       & 1.8                            \\ \bottomrule
            \end{tabular}
    }
    \end{center}
    \vspace{-0.50cm}
    \caption{\textbf{The SGG performance and inference time with different length of output sequences.} \textbf{SL}: output sequence lengths; \textbf{Sec}: inference time of second per image. } 
    \vspace{-0.59cm}
    \label{tab:perform_seqlen}
\end{table}

\vspace{-0.1cm}
\subsection{Model Analysis}
\vspace{-0.1cm}

In this section, we analyze our sequence generation-based SGG framework.
We explore the impact of different prefix prompts on scene graph generation and evaluate the diversity, quality, and time consumption of generated scene graph sequences. We also present the visualization of generated sequences in supplementary.

\noindent \textbf{Quality of Scene Graph Sequence}
We use a heuristic rule to parse the triplet sequence based on the scene graph prompt.
To showcase our approach's effectiveness, we track [REL] token occurrences and the number of relation triplets across different sequence lengths, as shown in Tab.~\ref{tab:abl_valtrp}.
The results demonstrate that VLM generates the formatted scene graph sequence, with the heuristic rule efficiently retrieving most relation triplets.

\begin{table}[]
    \begin{center}
    \resizebox{0.37\textwidth}{!}{ 
        \begin{tabular}{l|ccc|c}
            \toprule
            \textbf{SL.} & \textbf{\#Trip.} & \textbf{\#Uni.Trip.} & \textbf{\#[REL]} & \textbf{\% Valid}  \\ \midrule
            1024 & 87.2  & 53.4 & 95.1     & 95.3\%               \\
            768 & 76.1  & 40.2 & 79.2     & 96.0\%               \\
            512 & 45.0  & 34.3 & 46.2     & \textbf{97.4\%}               \\
            256 & 22.4  & 20.6 & 23.3     & 96.1\%               \\
            128 & 17.4  & 11.3 & 18.7     & 93.0\%               \\ \bottomrule
            \end{tabular}
    }
    \end{center}
    \vspace{-0.47cm}
    \caption{\textbf{Analysis for quality of generated Scene graph sequences.} 
    SL.: Generated sequence Length; 
    \# Trip.: Average of number of relation triplets;
    \# U. Trip.: Average of number of unique relation triplets;
    \# [REL]: Average of number of occurrence of  [REL] tokens;
    \% Valid: percentage of valid relation triplets. } 
    \label{tab:abl_valtrp}
    \vspace{-0.32cm}
\end{table}

\noindent \textbf{Time Consumption of Sequence Generation}
Our framework also achieves a good trade-off between performance and computational cost by varying the lengths of the generated sequences.
As shown in Tab.~\ref{tab:perform_seqlen}, \MT{} maintains comparable performance when the output length is reduced by 50\%, with comparable inference times to 2-stage SGG methods (more comparison is in supplementary).

\begin{table}[]
    \begin{center}
    \resizebox{0.42\textwidth}{!}{ 
        \begin{tabular}{l|ccc|ccc|c}
            \toprule
            \multirow{2}{*}{\textbf{Model}}   & \multicolumn{3}{c|}{\textbf{RC}} & \multicolumn{3}{c|}{\textbf{RC+}} & \textbf{RCg} \\ \cmidrule{2-8} 
                & val      & testA     & testB     & val       & testA     & testB     & val          \\ \midrule
            init  & 83.1     & 86.0     & 77.0     & 77.1      & 83.1      & 69.8      & 74.3         \\
            \textbf{\MT{}}  & \textbf{86.0}     & \textbf{88.9}     & \textbf{82.4  }    & \textbf{79.8}      & \textbf{84.3 }     & \textbf{72.3}      & \textbf{77.8}         \\ \bottomrule
        \end{tabular}
    }
\end{center}
\vspace{-0.55cm}
\caption{\textbf{Performance of visual grounding task.} \textbf{RC}, \textbf{RC+} and \textbf{RCg} represents the RefCOCO, RefCOCO+ and RefCOCOg datasets respectively. } 
\vspace{-0.65cm}
\label{tab:visual_grounding}
\end{table}

\vspace{-0.06cm}
\subsection{Downstream VL-Tasks}
\vspace{-0.05cm}
We apply \MT{} to VL tasks, including visual grounding, visual question answering, and image captioning. 
Through comparisons with the initial pre-trained model, we assess how explicit scene representation modeling influences vision-language reasoning.

\noindent \textbf{Visual Question Answering}
On the GQA benchmark, \MT{} outperforms the initial pre-trained model by 1.7 in overall accuracy, with the largest gains in relation (1.9) and object (1.3) questions. 
Additionally, our unified SGG training also boosts BLIPv2 VLM's zero-shot performance on GQA from 32.3 to 33.9, with improvements across all question types.

\begin{table}[]
    \begin{center}
    \resizebox{0.5\textwidth}{!}{ 
    \begin{tabular}{l|cccccc}
    \toprule
    \multirow{2}{*}{\textbf{Model}} & \multicolumn{6}{c}{\textbf{Question Type}}                                                                        \\ \cmidrule{2-7} 
       & \textbf{Relation} & \textbf{Attribute} & \textbf{Object} & \textbf{Category} & \textbf{Global} & \textbf{Overall} \\ \midrule
    BLIP & \multicolumn{6}{c}{Finetuned} \\ \midrule
    Init   & 54.9    & 64.5      & 86.6   & 62.2   & 67.6            & 62.5             \\
    \textbf{\MT{}}-E   & 55.8    & 64.3   & 87.5   & 62.6    & 67.9  & 63.1    \\ 
    \textbf{\MT{}}   & \textbf{56.8}     & \textbf{66.3}      & \textbf{87.9}   & \textbf{62.8}     & \textbf{68.1}   & \textbf{64.2}    \\ \midrule
    BLIPv2 & \multicolumn{6}{c}{Zeroshot} \\ \midrule
    Init   & \textbf{29.8}     & \textbf{28.8}      & \textbf{53.6}   & \textbf{31.1}     & \textbf{28.2}   & \textbf{32.3}    \\  
    \textbf{\MT{}}  & \textbf{31.3}     & \textbf{29.6}      & \textbf{53.9}   & \textbf{34.3}     & \textbf{26.1}   & \textbf{33.9}    \\ \bottomrule
    \end{tabular}
    }
    \end{center}
    \vspace{-0.53cm}
    \caption{\textbf{Performance comparison between initial pre-trained VLM and \MT{} on GQA).} We report the answering accuracy respectively according to the question type proposed by benchmark.
    E: denotes the entity-only scene graph sequence for ablation; 
    }
    \vspace{-0.35cm}
     \label{tab:GQA}
\end{table}

\begin{table}[]    
    \begin{center}
        \resizebox{0.34\textwidth}{!}{ 
        \begin{tabular}{l|l|ccc}
        \toprule
        &\textbf{}             & \textbf{Bleu4} & \textbf{CIDEr} & \textbf{SPICE} \\ \midrule
        \multirow{3}{*}{BLIP}&    Init   & \textbf{0.44}           & \textbf{132.1}          & 23.6           \\
        &Det-T             & 0.43           & 132.0          & 23.5           \\
        &\textbf{\MT{}}    & 0.41           & 131.2          & \textbf{23.9}           \\ \midrule
        \multirow{2}{*}{BLIP v2 } &Init               & 0.42           & 145.1           & 25.6           \\
        &\textbf{\MT{}} & \textbf{0.43}           & \textbf{145.2}          & \textbf{26.0}           \\ \midrule
        \end{tabular}
        }
        \vspace{-0.21cm}
        \caption{\textbf{Performance comparison on COCO image captioning.}  }
        \vspace{-0.95cm}
    \end{center}
    \end{table}

\noindent \textbf{Image Captioning}
We conduct the evolution of image captioning on COCO~\cite{chen2015microsoft}.
The results demonstrate that our framework achieves comparable performance to the initial pre-trained VLMs, including BLIP and BLIPv2, especially the SPICE metric~\cite{anderson2016spice}, which indicates the scene description quality of the generated caption.

\noindent \textbf{Visual Grounding}
Our method surpassed initial pre-trained model by 2.9 and 5.4 on the validation and test A/B splits of RefCOCO and by 2.7, 1.2, and 2.5 for RefCOCO+ and RefCOCOg, respectively (see Tab.~\ref{tab:visual_grounding}). These results underscore the effectiveness of knowledge transfer in SGG modeling, enhancing reasoning and localization following unified SGG training for visual grounding.

\vspace{-0.2cm}
\section{Conclusion}
\vspace{-0.2cm}

In this paper, we present a VLM-based open-vocabulary scene graph generation model that formulates the SGG through an image-to-graph translation paradigm, thereby addressing the difficult problem of generating open-vocabulary predicate SGG.
The holistic architecture unifies the SGG and subsequent VL tasks, thereby enabling explicit relation modeling to benefit the VL reasoning tasks.
Extensive experimental results validate the validity of the proposed model.

\noindent \textbf{Discussion of Limitations}:

\noindent \textit{Performance on Close-vocabulary SGG.}
Due to the inherent weakness of the visual backbone and labeling noise, our \MT{} performs suboptimal on VG and PSG datasets in a standard setting.
Improving the perception performance of VLMs with high resolution input is a challenging and important issue for future research.

\noindent \textit{Study on More VL Models and Tasks.}
Furthermore, our study primarily focuses on the BLIP applied to the SGG and BLIP and BLIPv2 for VL tasks: VQA, visual grounding, and image captioning.
However, our approach exhibits potential for extension to other VL models and tasks.
Exploring these possibilities remains for future research.

\noindent \textbf{Acknowledgement}:
This work was supported by NSFC under Grant 62350610269, National Key R\&D Program of China 2022ZD0161600, Shanghai Frontiers Science Center of Human-centered Artificial Intelligence, MoE Key Lab of Intelligent Perception and Human-Machine Collaboration (ShanghaiTech University), 
and Shanghai Postdoctoral Excellence Program 2022235.

{
    \small
    \bibliographystyle{ieeenat_fullname}
    \bibliography{main}
}

\newpage
\newpage
\appendix

\section{More Experiment Configurations}

\paragraph{Datasets and tasks}
We evaluate our method on both SGG task and downstream VL-tasks.
For SGG task, we use two large-scale SGG benchmarks: Panoptic Scene Graph Generation~(PSG)~\cite{yang2022panoptic}, Visual Genome~(VG)~\cite{krishna2017visual}. 
We mainly adopt the data splits from the previous work~\cite{zellers_neural_2017, li2021bipartite, yang2022panoptic}.
For Visual Genome~\cite{krishna2017visual} dataset, we take the same split protocol as~\cite{xu_scene_2017,zellers_neural_2017}
where 62,723 images are used for training, 26,446 for test, and 5,000 images sampled from the training set for validation. 
The most frequent 150 object categories and 50 predicates are adopted for evaluation.
The Panoptic Scene Graph Generation~\cite{yang2022panoptic} 
has 4,4967 images are used for training, 1,000 for test, and 3,000 images sampled from the training set for validation. 
There 133 object categories and 56 predicates categories in total.
We use it bound box annotation for SGG task rather than segmentation masks.

For the open-vocabulary predicate SGG settings, we randomly select 30\% predicate categories as novel class. 
For VL tasks, we inspect our model on VL-task which potentially need the visual scene representation, such as visual grounding on RefCOCO/+/g~\cite{yu2016modeling, mao2016generation}, visual question answering on GQA~\cite{hudson2019gqa}, and image captioning on COCO image caption~\cite{chen2015microsoft}.

\paragraph{Implementation Details}
We initialize our PGSG by using the BLIP \cite{li2022blip} model with ViT-B/16 as the visual backbone and BERT$_{base}$ as the text decoder.
For scene graph training process, we use the image size of 384 times 384, an AdamW \cite{loshchilov2017decoupled} optimizer with lr = 1e-5, weight decay of 0.02 with a cosine scheduler.We increase the learning rate of position adaptors to 1e-4 for faster convergence.
We train our model on 4 A100 GPUs with 50 epoch.
For downstream tasks fine-tuning, we following the training setup of BLIP \cite{li2022blip}. 
We use the image encoder and text decoder of PGSG model, and the text encoder and word embedding remain the same as in the pre-trained BLIP model.
During the scene graph sequence generation, we generate M=32 number of sequences which length is L=24. 
For category amplifier $\beta_i$, we set this hyper-parameter as 5.0 for entity categories and 1.0 for predicate classification.

\section{More Experimental Results}
In this section, we propose mre Experimental results, includes quantitative and quantitative analysis of our method.

\section{Quantitative Results}
\begin{table}[t]
    \begin{center}
        \resizebox{0.47\textwidth}{!}{ 
            \begin{tabular}{l|l|c|cccc|c}
                \toprule
                \multirow{3}{*}{\textbf{B}} & \multirow{3}{*}{\textbf{M}} & \multirow{2}{*}{\textbf{Zs Trp.}} & \multicolumn{5}{c}{\textbf{Standard SGG}}                                                                                              \\
                                            &                             &                                          & \multirow{2}{*}{mR50} & \multicolumn{1}{c|}{\multirow{2}{*}{R50}} & \multicolumn{2}{c|}{wmAP}            & \multirow{2}{*}{score\_wtd} \\
                                            &                             & R@50/100                                 &                       & \multicolumn{1}{c|}{}                     & phr      & \multicolumn{1}{c|}{rel}  &                             \\ \midrule
                \multirow{3}{*}{R101}       & RelDN                       & -                                        & 39.7                  & \multicolumn{1}{c|}{72.1}                 & 28.7     & \multicolumn{1}{c|}{29.1} & 38.6                        \\
                                            & HOTR                        & -                                        & 36.8                  & \multicolumn{1}{c|}{52.6}                 & 21.5 & 19.4                  & 26.8                        \\
                                            & SGTR                        & -                                        & 38.6                  & \multicolumn{1}{c|}{59.1}                 & 36.9     & \multicolumn{1}{c|}{38.7} & 42.8                        \\ \midrule
                \multirow{2}{*}{ViT-B*} & SGTR                        & 19.4/31.6                                & 30.5                  & \multicolumn{1}{c|}{52.6}                 & \textbf{28.0}     & \multicolumn{1}{c|}{\textbf{22.7}} & \textbf{30.8}                        \\
                                            & \textbf{\MT{}}      & \textbf{23.1}/\textbf{38.6 }                 & \textbf{40.7}         & \multicolumn{1}{c|}{\textbf{62.0} }       & 27.8     & \multicolumn{1}{c|}{19.7} & 28.7                        \\ \bottomrule
            \end{tabular}
        }
    \end{center}
    \caption{\textbf{The close-vocabulary SG-Det performance on OpenImage V6.} } \label{tab:stdsgg_oiv6}

\end{table}

\begin{table}[]
    \begin{center}
    \resizebox{0.48\textwidth}{!}{ 
        \begin{tabular}{l|ccc|c}
            \hline
            \multirow{3}{*}{\textbf{Prompt}} & \multicolumn{3}{c|}{\textbf{Open Vocab SGG} }    & \multicolumn{1}{c}{\multirow{2}{*}{\textbf{ZS Trp.}}}      \\
                                                 & \multicolumn{2}{c|}{\textbf{Novel+Base} }    & \textbf{Novel} \\
                                                 & mR@50/100    & \multicolumn{1}{c|}{R@50/100}         & mR@50/100        & R@50/100              \\ \midrule
            A   & 8.2/10.5         & \multicolumn{1}{c|}{14.5/16.4}          & 2.3/7.0          & 3.6/6.4           \\
            B   & \textbf{9.3/11.7}  & \multicolumn{1}{c|}{\textbf{17.7/20.4}} & 3.7/8.6          & \textbf{4.4/7.6}  \\
            C  & 9.1/10.1          & \multicolumn{1}{c|}{16.3/19.4}          & \textbf{4.1/9.0} & 4.1/6.6           \\ \midrule
        \end{tabular}
    }
    \end{center}
    \caption{\textbf{Ablation study on different prompt for SGG task on PSG dataset.} 
    A: "A visual scene of: "; B: "Describe the image by relationships:"; C: "A picture of: "  } 
    \label{tab:Ablation_prmpt}
\end{table}

\begin{figure*}[h]
    \centering
    \includegraphics[width=0.99\linewidth]{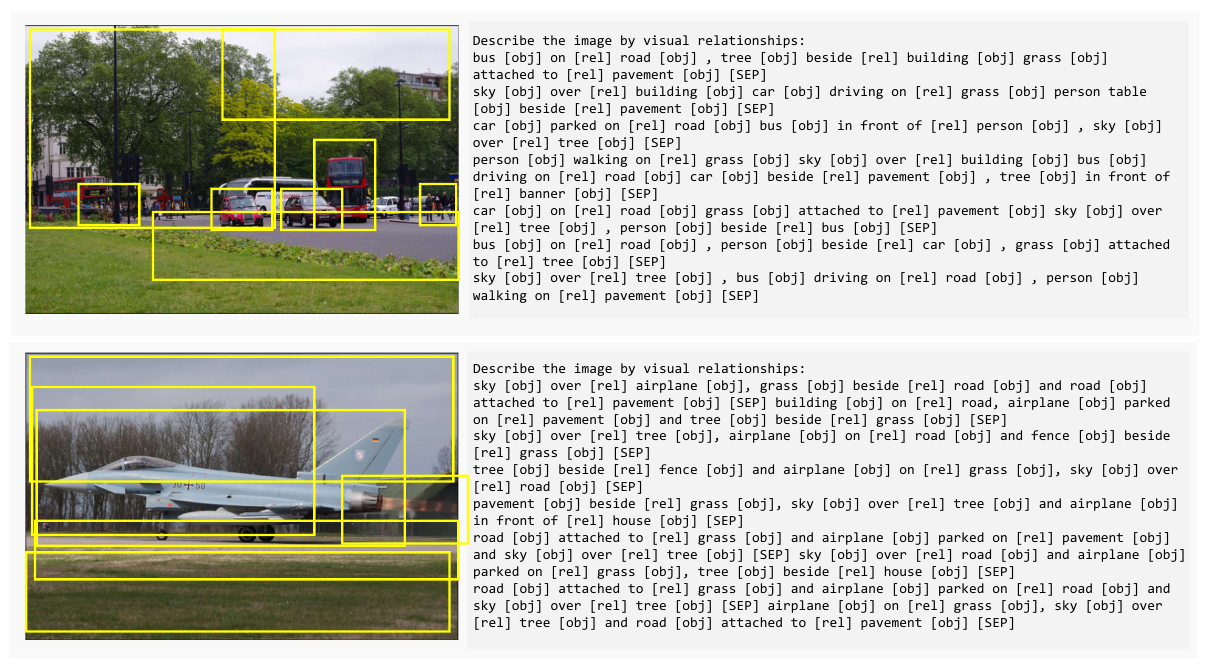}
    \caption{\textbf{The visualization of scene graph sequence prediction of PGSG.}}
    \label{fig:vis}
\end{figure*}

\subsection{Close-vocabulary SGG on OpenImage V6}
In Tab.~\ref{tab:stdsgg_oiv6}, we present the close-vocabulary SG-Det performance on OpenImage V6 across various visual backbones and zero-shot triplet (\textbf{Zs Trp.}) scenarios.
With the same backbone as BLIP ViT-B, our PGSG achieves comparable performance with baseline SGTR in a standard close-vocabulary setting and reasonable performance with the previous one-stage SGG method, which has a larger input resolution ResNet-101 backbone.
For compositional generalization setting, zero-shot triplet SGG, our method achieves a remarkable 7.0 improvement over the SGTR baseline.

\subsection{Sensitivity of different Prefix Prompts} 
We also study the different prompt structures for generating the scene graph, as shown in Tab.~\ref{tab:Ablation_prmpt}.
We experiment with the PGSG framework with different prefix instructions for the scene graph generation task.
The results show that more specific instructions yield a slight improvement in performance, which indicates that our method has robustness for different instructions.

\subsection{Time Complexity Comparison With Previous Method} 
Despite potential inference time increases due to the self-regression generation with a large model, we have effectively mitigated this issue by reducing output size.
We achieve a boost in inference speed reasonable open-vocabulary SGG performance, in the Tab.~4 of the main paper.
We also compare the inference times with other SGG methods, as shown in Tab.~\ref{tab:time-cplx}.  The results demonstrate that PGSG attains comparable time efficiency while maintaining its competitive open-vocabulary SGG performance.

\subsection{Time Complexity}
Despite potential inference time increases due to the self-regression generation with a large model, we have effectively mitigated this issue by reducing output size.
We achieve a boost in inference speed reasonable open-vocabulary SGG performance, in the Tab.~4 of the main paper.
We also compare the inference times with other SGG methods, as shown in Tab.~\ref{tab:time-cplx}.  The results demonstrate that PGSG attains comparable time efficiency while maintaining its competitive open-vocabulary SGG performance.

\begin{table}[t]
    \begin{center}
        \resizebox{0.42\textwidth}{!}{ 
          \begin{tabular}{l|ccccc}
            \toprule
            \textbf{M}    & VCTree & GPS-Net & BGNN & PGSG & PGSG* \\ \midrule
            \textbf{Time} & 1.69   & 1.02    & 1.32 & 4.8  & 1.8   \\ \bottomrule
            \end{tabular}
        } 
    \end{center}
    \caption{\textbf{the inference speed (Second per image) comparison with previous two-stage SGG methods} } \label{tab:time-cplx}
  
\end{table}


\section{Qualitative Results}
We also present the qualitative analysis for the PGSG framework to take a close look at the sequence generation-based SGG framework.
In Fig.~\ref{fig:vis}, we show a few examples of generated sequences from our validation set of the PSG dataset.
At inference time, the VLM generates scene graph sequences with entity-aware tokens as indicators by using several short token sequences with nucleus sampling, which are able to obtain diverse visual relations.
The following entity grounding module extracts the boxes for each entity within the sequences.

\end{document}